\documentclass[sigconf]{acmart}

\usepackage{booktabs} 
\usepackage{subfig}

\setcopyright{rightsretained}

\acmDOI{10.475/123_4}

\acmISBN{123-4567-24-567/08/06}

\acmArticle{4}
\acmPrice{15.00}

\begin{document}
\title{Mammography Dual View Mass Correspondence}

\author{Shaked Perek}
\affiliation{%
  \institution{IBM Research - Haifa, Israel}
}
\email{shaked.perek@il.ibm.com}

\author{Alon Hazan}
\affiliation{%
  \institution{IBM Research - Haifa, Israel}
}
\email{alon.hazan@il.ibm.com}

\author{Ella Barkan}
\affiliation{%
  \institution{IBM Research - Haifa, Israel}
}
\email{ella@il.ibm.com}

\author{Ayelet Akselrod-Ballin}
\affiliation{%
  \institution{IBM Research - Haifa, Israel}
}
\email{ayeletb@il.ibm.com}


\begin{abstract}
Standard breast cancer screening involves the acquisition of two mammography X-ray projections for each breast. Typically, a comparison of both views supports the challenging task of tumor detection and localization. We introduce a deep learning, patch-based Siamese network for lesion matching in dual-view mammography. Our locally-fitted approach generates a joint patch pair representation and comparison with a shared configuration between the two views. We performed a comprehensive set of experiments with the network on standard datasets, among them the large Digital Database for Screening Mammography (DDSM). We analyzed the effect of transfer learning with the network between different types of datasets and compared the network-based matching to using Euclidean distance by template matching. Finally, we evaluated the contribution of the matching network in a full detection pipeline. Experimental results demonstrate the promise of improved detection accuracy using our approach.
\end{abstract}

%
%


\keywords{Biomedical Imaging, Deep learning, Mammography}

\maketitle

\section{Introduction}
Mammography (MG) the primary imaging modality for breast cancer screening, typically utilizes a standard dual-view procedure. Two X-ray projection views are acquired for each breast, a craniocaudal (CC) and a mediolateral oblique (MLO) view. Breast cancer abnormalities include categories such as calcifications, architectural distortions asymmetries and masses \cite{giger2013breast}. Examining the correspondence of a suspected finding in two separate compression views, enables the radiologist to better classify an abnormality. Studies have shown that using a two-view analysis helps radiologists reduce false positive masses caused by overlapping tissues that resemble a mass, and ultimately helps achieve a higher detection rate \cite{warren1996value}. Although Computer Aided Diagnosis (CAD) algorithms were developed in the last two decades to assist radiologists, their usefulness has been debated. This is partially due to the many false positives they produce, especially for masses and architectural distortions. We propose a novel approach for identifying the correspondences between masses detected in different views, to further improve the detection and classification of MG algorithms.   

Previous work on MG classification employed hand-crafted features, such as texture, size, histogram matching, distance from the nipple, and more. The extracted features were then classified together using various techniques to assess the similarity between image pairs. \cite{paquerault2002improvement} Demonstrated the positive effect of dual-view analysis, which detects suspicious mass in one view and its counterpart in the other view. Based on geometric location, this analysis fuses both sets of features and classifies them with linear discriminant analysis. \cite{amit2015automatic} used dual view analysis to improve single-view detection and classification performance by combining the dual-view score with the single-view score. The dual view features were obtained manually using candidate location, shape, and image characteristics.

Deep learning approaches have already shown impressive results in MG detection and classification. Bekker et al. \cite{bekker2016multi} present a micro-calcification classification approach that uses a dual-view method based on two neural networks; this is followed by a single neuron layer that produces the decision based on the concatenated features from both full image views. \cite{teare2017malignancy} Present a multiscale convolutional neural networks (CNN) for malignancy classification of full images and sub-image patches integrated with a random forest gating network.

Dhungelz et al. \cite{dhungel2017fully} proposed a multi-view deep residual network (Resnet) to automatically classify MG as normal/benign/ malignant. The network consists of six input images, CC and MLO together with binary segmentations of masses and micro-calcifications. The second-to-last layer concatenates the output of each Resnet, followed by a fully connected layer that determines the class. Similarly, \cite{geras2017high} conducted a study on a large dataset and proposed a two-stage network approach that operates on the four full images: CC and MLO of the left and right breasts. The second stage concatenates the four view-specific representations to a second softmax layer, producing the output distribution.
 
 Most multi-view deep learning approaches to MG are applied on unregistered full images and concatenate the features obtained by the network on each view separately. In contrast, we propose a Siamese approach that focuses on matching localized patch pairs of masses from dual views. Siamese networks are neural networks that contain at least two sub-networks, with identical configuration, parameters, and weights. During training, updates to either path are shared between the two paths. To address the correspondence problem, previous works used the \textbf{Siamese network} \cite{koch2015siamese} to simultaneously train inputs together. \cite{chopra2005learning} this type of network for a face verification task, in which each new face image was compared with a previously known face image. \cite{wang2017multi} Demonstrate the advantage of Siamese networks by detecting spinal cord mass in different resolutions. Sharing parameters leads to fewer parameters allowing training with smaller datasets. The subnetworks representation is related, and thus better suited for the comparison task.

Our work entails three key contributions: 1) A novel deep learning dual view algorithm for mass detection and localization in breast MG based on Siamese networks, which have not been used before to solve lesion correspondence in MG. 2) A careful set of experiments using several datasets to study the contribution of the network components, also showing that the network is better than the classic template matching approach. 3) Evaluation of our approach on the DDSM database.


\section{Methods}
\label{sec:2}
For this study, our input took unregistered CC/MLO MG images and matched between lesions appearance in both views. Below, we describe the network matching architecture, the experimental methodology including fine-tuning and comparison to template matching and how the matching architecture is integrated into an automatic detection pipeline.

\subsection{Matching Architecture}
\label{sub:A}
Our approach extends the work presented by Han et al.\cite{han2015matchnet}. The authors developed MatchNet, a CNN approach for patch-based matching between two images. The network consists of two sub-networks. The first is a \textbf{feature network}, a Siamese neural network, in which a pair of patches, extracted from the CC and MLO views are inserted and processed through one of two networks. Both paths consist of interchanging layers of convolutions and pooling, which are connected via shared weights. The second is the \textbf{metric network}, which concatenates the two features, contains three fully connected layers and uses a softmax for feature comparison. Dropout layers were added after layers FC1 and FC2 with value of 0.5. The network is jointly trained with a cross entropy loss. Figure 1, presents the modified network, including the networks ensemble approach.

The mammography datasets employed for this study were created by defining a positive image pair label, as the detections annotated by a radiologist in each view, while a negative pair label is defined by matching false detection with annotated detections in the other view.

\begin{figure}
\includegraphics[scale=0.19]{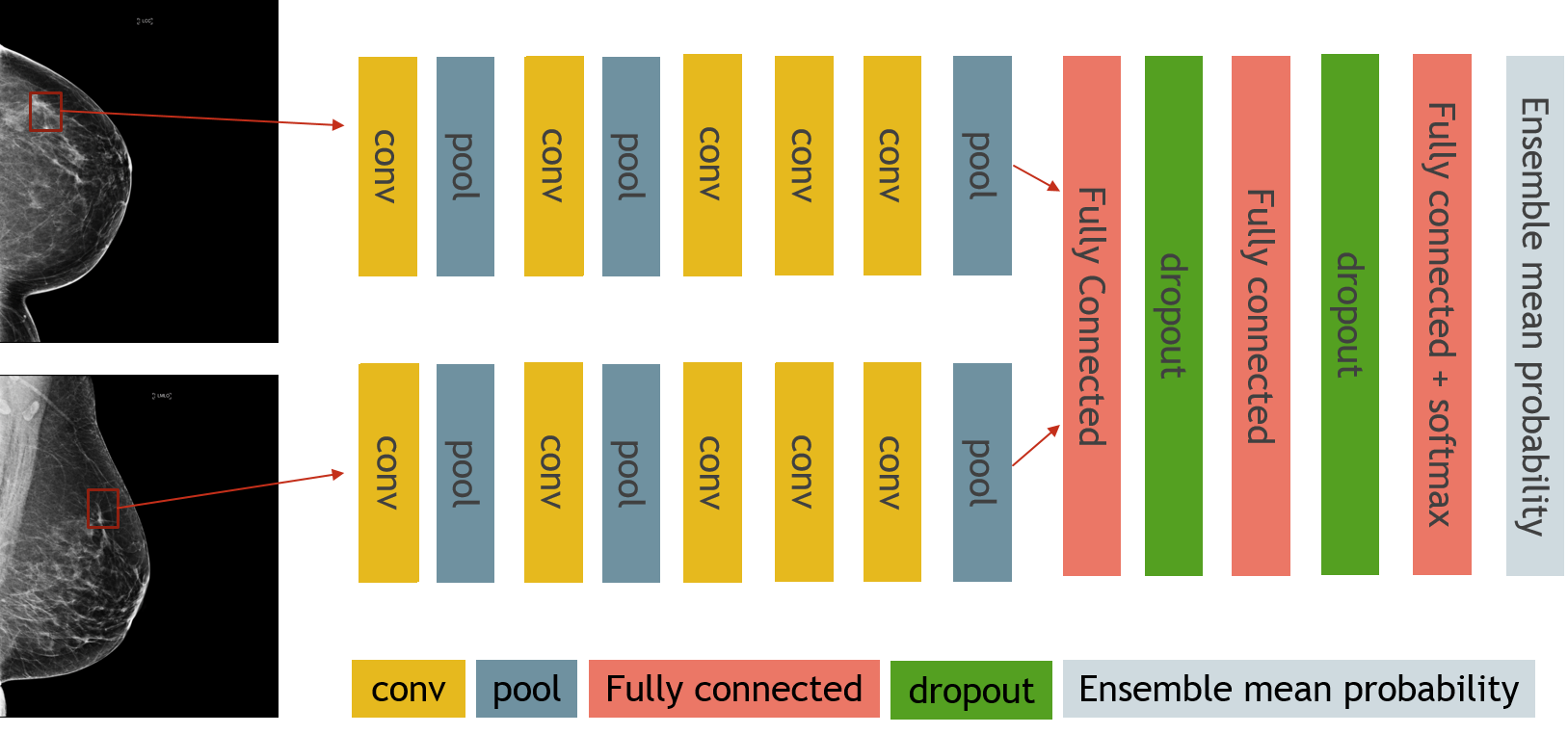}
\caption{The dual-view matching architecture. Patch pairs from CC and MLO views are inserted to the network. The feature network, consists of interchanging layers of convolutions and pooling, share parameters between paths. The metric network has fully connected layers with dropout, produce the final decision by networks ensemble.}
\end{figure}



\subsection{Fine Tuning the Network}
\label{sub:B}
Fine-tuning and transfer learning have shown to improve performance results despite of specific application domains \cite{tajbakhsh2016convolutional,yosinski2014transferable}. To adapt MatchNet to the task of matching detections from different MG views, we first evaluated fine-tuning. We fine-tuned by training the layers of the metric network, i.e. the three fully connected layers and the last convolution layer from the feature network. We used three different datasets, as described in the Experiment and Results section, including: Photo tourism (natural image pairs)\cite{data}, Digital Database for Screening Mammography (DDSM) \cite{heath2000digital} and In-house dataset. We used the trained weights of one dataset domain to fine tune the other datasets.

\subsection{Template Matching}
\label{sec:C}
Template matching, which extracts sub-image patches and computes a similarity measure that reflects the template and image patch correspondence, has been used extensively in computer vision \cite{Ballard:1982:CV:578131}. We compare our deep learning network to template matching with normalized cross correlation. Intuitively, we assume that the similarity of image patches of a mass in one view with the same mass in the other view under deformations, will be higher than the similarity with a different mass or region of the breast \cite{giger2013breast}.

\subsection{Multi-view Automatic Lesion Detection}
\label{sub:D}
We integrated two components, a \textbf{matching architecture} and a single-view \textbf{detection algorithm} to exploit the contribution of the dual-view network to the full pipeline. The detection algorithm is based on a modified version of U-net \cite{ronneberger2015u}, which was originally designed for the biomedical image processing field. In the original U-net, the output size is identical to the input size. However, for our task segmentation is not required at the pixel level, since the boundary of tumors and healthy tissue is ill-defined. Thus, we modified the U-net output, so that each pixel of the output, corresponds to a 16x16 pixels area of the input. 

The system flow is such that, given a dual-view pair of images as input, the single-view detection algorithm is applied separately on the CC, MLO image $I_{cc}, I_{MLO}$ and outputs a set of candidate patches, $P_{CC}=\{p_{CC}^1,...p_{CC}^N\}, P_{MLO}=\{p_{MLO}^1,...p_{MLO}^M\}$ respectively. The objective of the matching architecture is to identify the correspondences. If both patch candidates, CC and MLO views, from the detection flow, are identified as a true lesion, than the label for the pair will be true and accordingly considered a positive match. We assign labels to each pair based on the Dice Coefficient threshold $\delta$, between two masks, defined by a detection contour and ground truth lesion contour respectively. For our experiments, we used $\delta =0.1$ as the threshold. Any contour with a larger score is said to be a true lesion.

\subsection{Ensemble}
\label{sub:E}
Medical datasets are generally unbalanced. Namely, the number of positive pairs are significantly smaller than the negative pairs. Thus, we train two networks, each network has a balanced input of positive pairs and randomly selected negative pairs. In the testing stage, we evaluate each test image through all networks, and achieve a final score using a mean probability.

We performed training with a learning rate of 0.0001, Adam optimizer and batch size of 512 without regularization. Experiments were performed on a Titan X Pascal GPU. The training time for DDSM models took 4 hours. The testing time on all detection pairs from same breast views, with model ensemble took 6 sec.

\begin{figure}[b]
\includegraphics[scale=0.6]{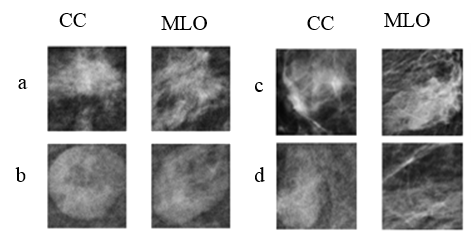}
\caption{Illustration of ROI input patches from two views, CC and MLO. (a,b) matching pairs, (c,d) non mathing pairs.}
\label{fig_pairs}
\end{figure}

\section{EXPERIMENTS AND RESULTS}
\subsection{Data Description}
\label{sub:3A}
We carried out the experiments on three different datasets: (a) The Photo Tourism dataset \cite{data}, consists of three image datasets: Trevi fountain, Notre Dame and Yosemite. Which is similar to the dataset used in the MatchNet paper \cite{han2015matchnet}. It consists of 1024 x 1024 bitmap images, containing a 16 x 16 array of image patches. Each image patch has 64x64 pixels and has several matching images that differ in contrast, brightness and translation. (b) The Digital Database for Screening Mammography (DDSM) \cite{heath2000digital}, contains 2620 cases of four-view mammography screenings. It includes radiologist ground truth annotations for normal, benign and malignant image. 1935 images contain tumors. (c) The In-house dataset includes benign and malignant tumor ground truth annotations, from both CC/MLO MG views for either left, right or both breasts. It contains 791 tumor pairs. Figure 2 shows some tumor pairs from In-house dataset used as positive examples for the network versus negative examples. We randomly split the data into training (80\%) and testing (20\%) subsets of patients. The partitioning was patient-wise to prevent training and testing on images of the same patient.

\begin{figure}
\includegraphics[scale=0.3]{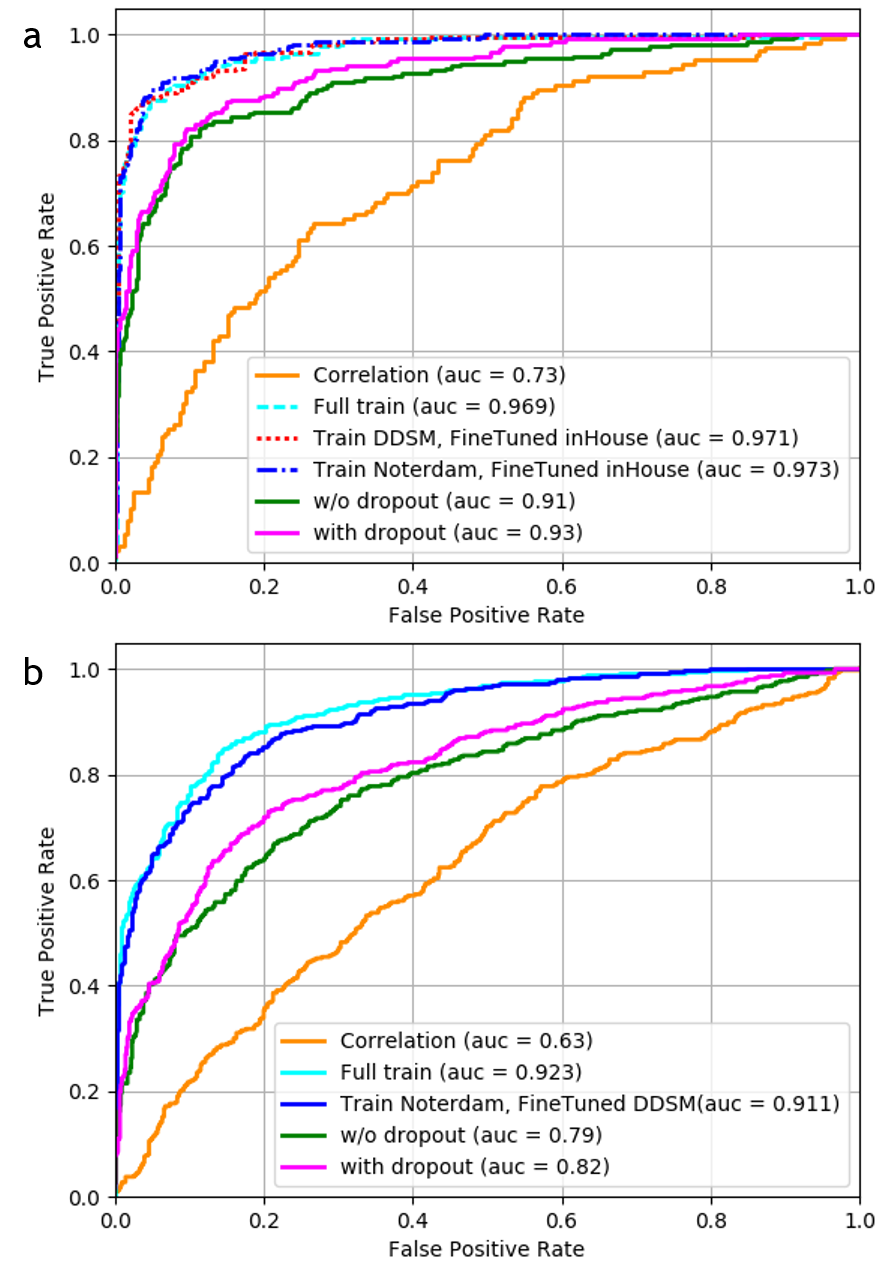}
\caption{Fine tuning ROC results. The figures demonstrate the different experiment performed to evaluate the ability of the matching architecture to classify MG pairs and non pairs. (a) In-house dataset shows no advantage for fine tuning. (b) DDSM dataset shows best result by full train (cyan).}
\label{fig_results}
\end{figure}

\subsection{Patch Preprocessing and Augmentations}
\label{sub:3B}
We extracted ROI patches from the full MG images of 4000x6000 pixels by cropping a bounding box around each detection contour. Each such bounding box was enlarged by 10\% in each dimension to include useful information around the lesion border. The extracted patches were then resized to 64x64 to generate the input to the network. We normalized all the datasets by subtracting the mean of each image and dividing by the standard deviation of each patch, avoiding the proposed MatchNet normalization [12].

Augmentation was utilized throughout the training stage on all three datasets, such that each patch was flipped left and right and rotated by 90°, 180°, 270°. Each augmented patch was matched with all the others augmented patches.

\subsection{Fine Tuning the Network}
\label{sub:3C}
We studied the contribution of fine-tuning on the results in three experiments. Full training on Photo tourism and fine tuning with (i) In-house (ii) DDSM (iii) Full training on DDSM and fine tuning with In-house.  (i+ii) were done using Notredam dataset. The results for these tests are presented in figure 3, where the upper and lower subfigures correspond to the In-house and DDSM dataset respectively. The comparison of the In-house and DDSM full training results (AUC 0.969, 0.92) with the fine tuning results (AUC 0.973,0.91) did not show a clear advantage over the fine tuning process. This can be explained by two factors: the domain transfer effect, namely despite of the Notredam large dataset of image pairs, natural images are different than medical images. Second, the Noterdam dataset pairs are much more similar to each other than the different views pairs from the breast images, which go through deformation.

Fine tuning the DDSM with the In-house dataset in (iii), obtained (AUC 0.971) compared to full training of (AUC 0.969). DDSM is a large MG dataset, however it is acquired with a different imaging technique from the In-house data (full field digital mammography) and this might explain the similar results. The ROC plot also shows the improvement in AUC by adding dropout in Figure 3.

\subsection{Template Matching}
\label{sub:3D}
The cross-correlation score was transformed from the range of [-1, 1] to the range of [0, 1] to represent the score as probabilities. The correlation presented in Figure 3 obtained significantly lower results of AUC 0.73, 0.63 on In-house, DDSM respectively.

\begin{figure}
\includegraphics[scale=0.35]{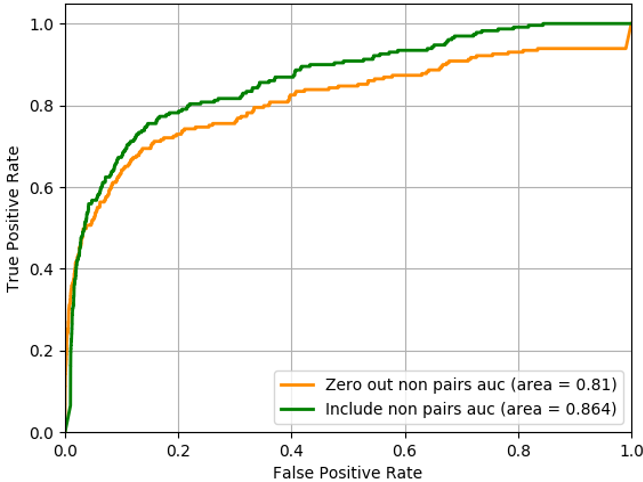}
\caption{Patch matching ROC using pipeline of automatic lesion detection. Green curve includes detections with no-pair in second view, orange curve excludes those detection.}
\label{fig_FP}
\end{figure}

\subsection{Multi-view Automatic Lesion Detection}
\label{sub:3E}
To evaluate the contribution of the matching architecture to the full detection pipeline, we applied the single-view detection algorithms on the CC, MLO image pairs followed by the matching architecture on the DDSM dataset. In some cases, detections will appear only for one view and not in the other. These cases cannot be evaluated using the matching architecture. Thus, two possibilities arise, to exclude all detections without a pair or to include them. Figure 4 shows the network classification of the set of patches into positive and negative matches, generates an AUROC of 0.864, 0.81 depending on whether the small set of detections with no-pairs were included or excluded. We conclude that it is reasonable to include these detection as some tumors may be identified only in a single view.

Additionally, the results in Figure 4 show that proposed approach can reduce the false positive detection rate while keeping a high sensitivity. Specifically, for the MG pairs matching, we can keep a sensitivity of 0.99 and specificity of 0.19. Namely, by keeping the standalone detections we are able to reduce the false positives by almost 20\%. Figure 5, illustrates the full pipeline prediction on the full MG images, where the probabilities of the false detections pairs (in cyan) are down-weighted and omitted in the final detection output. This approach is aligned with the approach used by human radiologists, first detecting suspicious findings and then analyzing them by comparing the multi-view appearance.

\begin{figure}
     \centering
     \subfloat[CC/MLO dual-view]{\includegraphics[scale=0.24]{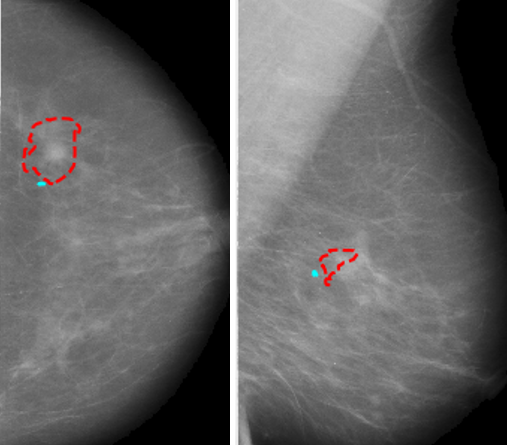}}
\subfloat[CC/MLO dual-view]{\includegraphics[scale=0.24]{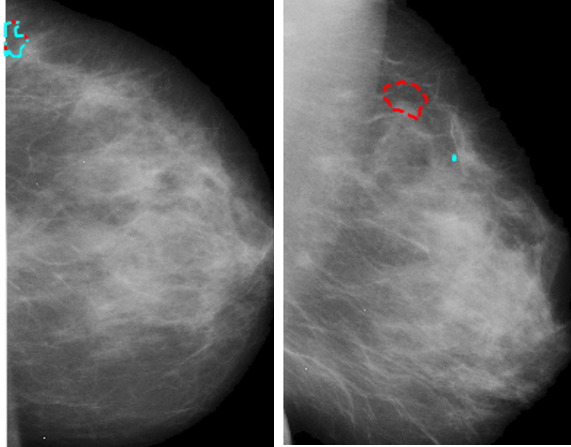}}
     \caption{Detection examples on DDSM dataset(a,b). Red contours denote automatically detected pairs that correspond to GT while, the cyan contours are false positive automatic detections that were reduced by the dual-view algorithm.}
     \label{det_pairs}
\end{figure}

\section{Discussion}
\label{sec:4}
Finding correspondence between patches from different views of the same breast is a challenging task. Each image from MLO/CC views undergoes nonlinear deformations which can make the lesions very different from each other. On the other hand, being able to detect the lesion in both views can help the radiologists reach more accurate findings. In this work, we propose a dual-view Siamese based network, in which the architecture learns a patch representation and similarity for lesion matching. We demonstrate the advantage of a learned distance metric implemented in the network and its value in addition to a single view detection. This work can also easily be extended to 3D mammography by applying 3D patches.  Future work will extend this work to other types of findings such as calcifications and will utilize mass location information to better eliminate false positives.

\bibliographystyle{ACM-Reference-Format}
\bibliography{sample-bibliography}

\end{document}